\documentclass[runningheads]{llncs}

\usepackage{eccv}

\usepackage{eccvabbrv}

\usepackage{graphicx}
\usepackage{booktabs}

\usepackage[accsupp]{axessibility}  

\usepackage{hyperref}

\usepackage{orcidlink}

\begin{document}

\title{Aligning Neuronal Coding of Dynamic Visual Scenes with Foundation Vision Models}

\titlerunning{Aligning Neuronal Coding with Foundation Vision Models}

\author{
    Rining Wu\inst{1, 2}\orcidlink{0000-0002-8834-9316} \and
    Feixiang Zhou\inst{3}\orcidlink{0000-0003-4939-9393} \and
    Ziwei Yin\inst{2}\orcidlink{0009-0007-8430-443X} \and
    Jian K. Liu\inst{1, 2}\orcidlink{0000-0002-5391-7213}
}

\authorrunning{R. Wu et al.}

\institute{
    University of Leeds \texttt{ ml20r2w@leeds.ac.uk } \and University of Birmingham \texttt{ \{z.yin.3, j.liu.22\}@bham.ac.uk } \and Lancaster University \texttt{f.zhou3@lancaster.ac.uk} }

\maketitle

\begin{abstract}
    Our brains represent the ever-changing environment with neurons in a highly dynamic fashion.
    The temporal features of visual pixels in dynamic natural scenes are entrapped in the neuronal responses of the retina. It is crucial to establish the intrinsic temporal relationship between visual pixels and neuronal responses.
    Recent foundation vision models have paved an advanced way of understanding image pixels.
    Yet, neuronal coding in the brain largely lacks a deep understanding of its alignment with pixels.
    Most previous studies employ static images or artificial videos derived from static images for emulating more real and complicated stimuli.
    Despite these simple scenarios effectively help to separate key factors influencing visual coding, complex temporal relationships receive no consideration.
    To decompose the temporal features of visual coding in natural scenes,
    here we propose Vi-ST, a \textbf{s}patio\textbf{t}emporal convolutional neural network fed with a self-supervised \textbf{Vi}sion Transformer (ViT) prior,
    aimed at unraveling the temporal-based encoding patterns of retinal neuronal populations.
    The model demonstrates robust predictive performance in generalization tests.
    Furthermore, through detailed ablation experiments, we demonstrate the significance of each temporal module.
    Furthermore, we introduce a visual coding evaluation metric designed to integrate temporal considerations and compare the impact of different numbers of neuronal populations on complementary coding.
    In conclusion, our proposed Vi-ST demonstrates a novel modeling framework for neuronal coding of dynamic visual scenes in the brain, effectively aligning our brain representation of video with neuronal activity. The code is available at \href{https://github.com/wurining/Vi-ST}{github.com/wurining/Vi-ST}.
\end{abstract}

\section{Introduction}
The neuronal coding of our neural system to visual stimuli constitutes a pivotal research area in computational neuroscience. When discussing the visual system, it typically involves components such as the retina, Lateral Geniculate Nucleus, Primary Visual Cortex, and more advanced visual cortices. The retina, in particular, comprises photoreceptors, horizontal cells, bipolar cells, amacrine cells, and ganglion cells. Retina ganglion cells serve as the output terminals of the retina, generating a series of neural response signals, namely spikes. The retina, as the initial component of the entire visual system, plays a crucial role in the conversion of light signals into bioelectric signals, representing a focal point of interest in neuronal coding of visual information in terms of pixels.

\begin{figure}[!t]
    \centering
    \includegraphics[width=1.0\textwidth]{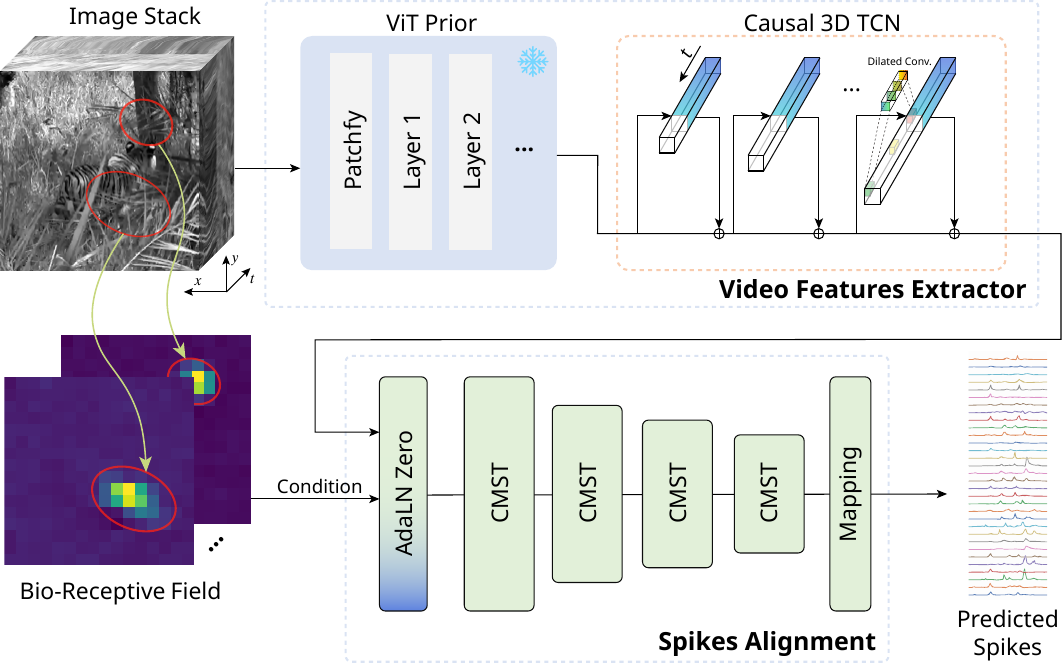}
    \caption{
        \textbf{The architecture of our proposed Vi-ST model.}
        The model comprises two primary components: a Video Feature Extractor based on
        the ViT prior, and a Spikes Alignment module. The original video is treated as
        a stack of frames, and spatial features are independently computed for each
        frame using pre-trained ViT. These features are packaged into a spatiotemporal
        cube of length $t$, which is subsequently processed by a Causal 3D Temporal
        Convolutional Network (C3TCN) to model spatiotemporal dynamics. The Receptive
        Field information of RGCs is utilized as a conditioning factor, fused with the
        spatiotemporal features through the 3D AdaLN Zero module, and subsequently fed
        into a series of Causal Multiscale Spatiotemporal (CMST) blocks. These CMST
        modules are employed for modeling the neuronal activity of a specific number of
        RGC neurons, ultimately yielding final spike predictions through a linear
        mapping. }\label{fig:model_arch}
\end{figure}

In experimental neuroscience, specific stimuli such as flashes, Gaussian noise, and enhanced static images are commonly used for controlled experiments to establish correlations between visual features and stimuli \cite{gollischEyeSmarterScientists2010}. However, real natural scenes are highly complex and dynamic, posing a significant challenge in accurately establishing the link between neural responses and scene changes \cite{liuSimpleModelEncoding2022, karamanlisRetinalEncodingNatural2022, liuInferenceNeuronalFunctional2017}. Recent advances, exemplified by recursive convolutional neural networks (RCNN)~\cite{zheng_unraveling_2021}, effectively elucidate the functioning of the recurrent circuitry in the biological retina. This suggests that visual stimuli in natural scenes can be decomposed into spatial and temporal domains, with temporal information being an indispensable component for understanding visual encoding. Nevertheless, efficiently modeling temporal or spatiotemporal information still requires further exploration.

Recently, in the field of deep learning, modeling methods based on temporal information have provided new insights for visual encoding~\cite{taylor_convolutional_2010,ji_3d_2013, carreira_quo_2018, wang_self-supervising_2021,koniusz_tensor_2022,farha_ms-tcn_2019}. When employing deep learning methods to address classic image or video tasks, it typically involves selecting local or global pixels and computing their correlations, ultimately mapping them to corresponding label domains. However, neuronal visual encoding tasks, in comparison, require the incorporation of more biologically plausible priors. This involves studying how neurons in the visual system respond to different visual features, patterns, and information, such as encoding for orientation, contrast, temporal characteristics, and the interaction between different neurons.

Furthermore, it is observed that neural responses often manifest as a series of highly dynamic and sparse irregular time sequences that require the spatiotemporal capabilities of deep learning models, together with video analysis~\cite{demb_functional_2015}. Therefore, the interesting and challenging question arises of how to effectively transfer powerful deep learning models to the task of neural visual encoding. In particular, previous work demonstrated that a properly designed model, such as RCNN~\cite{zheng_unraveling_2021}, is capable of learning the relatively simple and self-consistent video and their triggered neuronal response, e.g., training and testing data are taken from the same video where pixel context is conserved. However, these well-trained models cannot be transferred to other videos where the pixel complexities are quite different, not only for neurons in the retina but also for the visual cortex~\cite{turishchevaDynamicSensoriumCompetition2023a}.

To address these challenges, our work here introduces a spatiotemporal convolutional neural network (CNN) model~\cite{oord2016wavenet} based on the Vision Transformer (ViT) prior~\cite{dosovitskiy2021image}, termed Vi-ST. The ViT prior is derived from DINOv2~\cite{oquab_dinov2_2023}, which employs a self-supervised approach for pretaining on a huge dataset containing 1.2B unique images, yielding a set of adversarial feature representations with strong generality and generalization capabilities. Subsequently, we integrate the feature space of visual stimuli using a causal Temporal Convolutional Network (TCN) with 3D convolutional kernels. We incorporate receptive field information from retinal ganglion cells (RGCs) as auxiliary conditions. These, along with video features, are fed into a series of spatiotemporal convolutional modules aligning the voxel space of video spatiotemporal information with the neural response space of RGCs. In particular, we conducted ablation experiments, progressively removing specific modules from the baseline model to observe the correlation metrics of the predicted results. This demonstrates the effectiveness of the spatiotemporal modules.

Additionally, we find that employing a larger visual encoding space yields better performance in representing biological visual encoding. This is likely due to complementary encoding, where a too-small encoding space cannot capture highly dynamic and complex spikes effectively~\cite{ding_information_2023}.

Moreover, we introduce an indicator that considers the duration of neural responses to evaluate prediction performance. Specifically, we used kernel density estimation to fit density functions for the duration of spikes in two observation sequences. By comparing the similarity of these density functions, we assess the performance of the predicted results. In summary, our contributions can be summarized as follows:

\begin{itemize}
    \item Introducing Vi-ST, a spatiotemporal convolutional network with a pre-trained Vision Transformer (ViT) as a prior, designed for accurately establishing the mapping between complex dynamic natural scenes and retinal visual encoding.

    \item Analyzing the impact of complementary encoding on effectively representing neural responses by comparing predictions in different sizes of response spaces.

    \item Introducing an indicator that considers the duration of neuronal responses, providing a novel perspective for the analysis and observation of predicted neural encoding.
\end{itemize}

\section{Related Work}
In this section, we review the related work on retinal recurrent connection mechanisms, image and video-based methods, and 3D CNN-based temporal modeling.

\subsection{Retinal Recurrent Connection Mechanism}

The retina, being the starting point of the visual system, serves as an ideal model for the study of neural encoding and responses\cite{zapp_retinal_2022, zheng_unraveling_2021}. Located at the back of the eye, the retina consists of three layers of cells. Photoreceptor cells, including cones and rods, are responsible for brightness perception, respectively. Horizontal cells act as intermediary neurons connecting photoreceptor cells and bipolar cells, facilitating lateral inhibition to regulate information transfer between photoreceptors and bipolar cells, typically enhancing the retina's sensitivity to contrast and edges \cite{euler_retinal_2014,wu_two_2023}. Bipolar cells receive information from photoreceptor cells and transmit it to ganglion cells, regulating slight variations in light intensity and color. Amacrine cells are inhibitory interneurons that carry out inhibitory synapses, form complex visual processing, and have widespread gap junctions. Gap junctions between cells create electrical coupling, which strengthens synchronization between cells and regulates the activity of retinal cells\cite{pereda_gap_2013}. Ganglion cells are the final layer of neurons in the retina, transmitting processed signals to the optic nerve. Due to the lateral connections provided by gap junctions and amacrine cells, which serve as a mechanism akin to recurrent connectivity in the retina, this structure has been shown to model temporal information in visual stimuli more effectively than conventional CNNs \cite{zheng_unraveling_2021}.

\subsection{Image and Video Models}

\subsubsection{Self-supervised Foundational Vision Model}
With the advent of the Vision Transformer (ViT) model, many works have introduced the Transformer model into the image domain. The multi-head self-attention mechanism is a core component of the Transformer model \cite{vaswani_attention_2017}, which allows long-range context modeling and has been shown to be effective in capturing spatial information in images~\cite{dosovitskiy_image_2021}. DINOv2 is a foundational visual model trained using a discriminative self-supervised method, capable of obtaining robust frozen features without the need for fine-tuning\cite{oquab_dinov2_2023}. DINOv2 builds upon the foundation of DINO~\cite{caron_emerging_2021} and employs more complex and carefully crafted data augmentation strategies compared to its predecessor. DINOv2 has been proven to be a powerful universal visual feature extractor in various visual tasks, including medical image segmentation~\cite{anand_one-shot_2023}, video segmentation~\cite{zhang_dvis_2023}, pose estimation~\cite{chen_secondpose_2023}, 3D scene instance tracking~\cite{zhao_instance_2023}, dense point tracking~\cite{karaev_cotracker_2023}, and etc.

\subsubsection{3D CNN-based temporal modeling}

CNNs are specifically designed for image processing and computer vision tasks~\cite{bhatt_cnn_2021,lecun_gradient-based_1998}. The convolutional layers in CNNs, typically with 2 dimensional (2D) CNNs for vision tasks, can be regarded as filters with shared parameters at different locations, a concept common in biological visual systems\cite{bhatt_cnn_2021}. To incorporate temporal information into CNNs, 3D CNNs have been naturally extended from 2D CNNs, and have been widely used in video processing tasks~\cite{taylor_convolutional_2010,ji_3d_2013, carreira_quo_2018}. The Inflated 3D ConvNet (I3D), a widely used 3D CNN model, has demonstrated excellent performance in various video action recognition tasks and is widely utilized as a backbone for visual tasks involving temporal information, achieving impressive results~\cite{wang_self-supervising_2021,koniusz_tensor_2022,farha_ms-tcn_2019}. In particular, the Multi-Scale Temporal Convolutional Network (MSTCN) for video action recognition~\cite{farha_ms-tcn_2019}, which employs I3D as a backbone and utilizes dilated CNN modules to capture temporal information on different temporal scales. The Inception module of the I3D model effectively improves the performance of image tasks by integrating features from multiple branches at different scales. We continue to adopt the design philosophy of the Inception module to study the temporal relationship embedded in the neuronal coding of videos.


\section{Methodology}
In this section, we will provide a detailed explanation of the Vi-ST architecture and introduce the Vi-ST loss function. We divide the model into two parts: the Video Features Extractor and the Spikes Alignment module, as \cref{fig:model_arch}. In the feature extraction stage, we utilize the powerful foundation visual model, namely DINOv2, to be a prior for spatial information of image pixels.

\subsection{Video Features Extractor}

\begin{figure}[tbh]
    \centering
    \begin{subfigure}{0.48\linewidth}
        \includegraphics[width=1.0\linewidth]{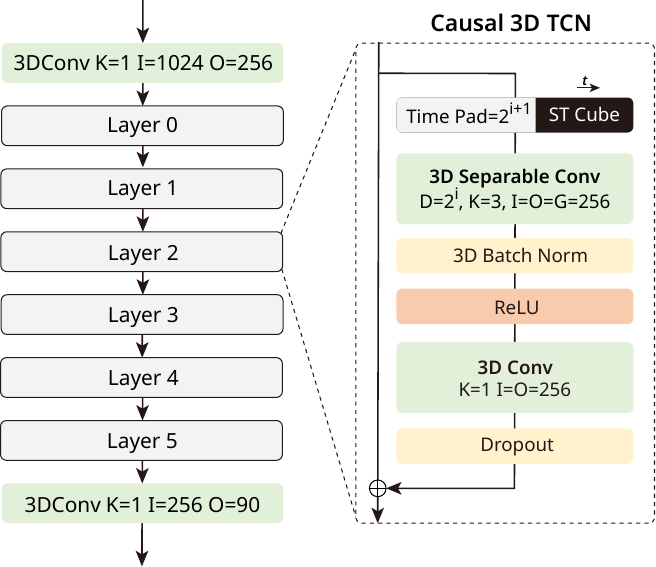}\caption{}\label{fig:video_features_extractor}
    \end{subfigure}
    \hfill
    \begin{subfigure}{0.48\linewidth}
        \includegraphics[width=1.0\linewidth]{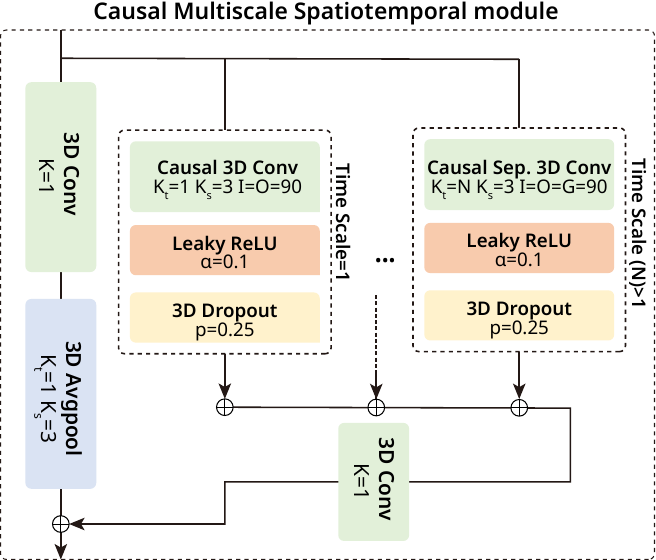}\caption{}\label{fig:spikest_alignment_module}
    \end{subfigure}
    \caption{
        \textbf{Vi-ST Submodules.}
        (a) The detailed architecture of the video features extractor.
        (b) The Causal Multiscale Spatiotemporal module (CMST) in the spikes alignment module.
    }\label{fig:vi_st_submodules}
\end{figure}

As \cref{fig:video_features_extractor} demonstrates, the video features extractor consists of multiple layers of C3TCN and 2 linear mapping layers. Where, the $D$ is the size of dilation, $K$ is the size of kernel, $I$ is the number of input channels, $O$ is the number of output channels, and $G$ is the number of filters' groups. The C3TCN is padding $2^{i+1}$ zeros for the $i$-th layer's input, along the time axis, and sets the dilation size to $2^{i}$. Meanwhile, the depth-wise separable convolution is used to reduce the number of parameters and mitigate overfitting. Ultimately, each layer is embedded with a residual connection.

Specifically, the Video Features Extractor is divided into two steps. Initially, the image stack $ X \in \mathbb{R}^{T\times H\times W \times C}$ is fed into the frozen DINOv2 model. DINOv2 transforms and ``patchfys'' each image into a sequence of patches $\epsilon \in \mathbb{R}^{T\times g \times g \times C}$. Subsequently, a reverse mapping is applied to $\epsilon$, restoring them to their original positions, resulting in $\epsilon \in \mathbb{R}^{T\times g\times g\times C}$. Since DINOv2 inherits the original ViT hierarchical structure\cite{dosovitskiy_image_2021}, which contains different semantics within different layers, we choose outputs of layer 0 to layer n as priors, where layer 0 represents patchify the image only.

In the second step, the prior is fed into the C3TCN to further model the spatiotemporal information of the video frames. C3TCN will map priors to $\epsilon \in \mathbb{R}^{T\times g\times g\times C'}$, where $C'$ is the number of RGCs. C3TCN extends the Residual 1D Dilated Convolution module of MSTCN by directly replacing the 1D CNN with Causal 3D CNN, maintaining MSTCN's approach of increasing receptive fields with layers. Here, C3TCN performs dilated convolution calculations only along the time dimension, while still keeping the spatial convolutional kernels size in 3$\times$3. Specifically, different time spans of receptive fields are set to capture diverse spatiotemporal features. Additionally, Batch Normalization\cite{ioffe_batch_2015} is added after C3TCN to prevent gradient explosions. This enhancement enables C3TCN to effectively and robustly capture spatiotemporal information in video data and provide comprehensive feature representations for subsequent modules.

\subsection{Spikes Alignment}

As \cref{fig:spikest_alignment_module} shows, the spikes alignment module mainly consists of the CMST, where the $D$ is the size of dilation, $K_t$ is the size of kernel along the time axis, $K_s$ is the size of kernel along the spatial axis, $I$ is the number of input channels, $O$ is the number of output channels, and $G$ is the number of filter groups.

For each CMST modules, different time scales are used to capture the multi-scale temporal features, and the features are added together, then fed into a mapping layer. Specifically, when the time scale is 1, the time kernel size is set to 1, and the spatial kernel size is kept to 3, then the vanilla causal 3D convolution is utilized. When the time scale is greater than 1, a causal separable 3D convolution with time kernel size $N$ and spatial kernel size 3 is used.

In the Spike Alignment module, we aimed to incorporate the RGC receptive field information to better align video pixel spaces and RGC spike spaces. Here, the receptive field, $RF \in \mathbb{R}^{W\times H}$, is downsampled along the spatial axis to match the size of $\epsilon \in \mathbb{R}^{T\times g\times g\times C'}$, and is used as a conditional factor for fusion with $\epsilon$ through the AdaLN Zero module\cite{peebles_scalable_2023}. As a result, we obtain the fused feature $F \in \mathbb{R}^{t\times g\times g\times C'}$. Since $\epsilon$ contains both temporal and spatial dimensions, we naturally extended AdaLN Zero to 3D AdaLN Zero.

Next, we feed the fused feature into the CMST block. CMST adds features from multiple temporal scale branches and maps them through a 1$\times$1$\times$1 3D CNN layer. Then, CMST utilizes residual connections to retain information from the previous block. To compress the spatial size of latent variables, we apply an average pooling operator on the previous layer's information, then add it to the outcome of multiscale branch. We set 4 CMST blocks in total, and the convolution kernel sizes of different time scales are shown in \cref{tab:kernel_size}.

\begin{table}[!bh]
    \caption{\textbf{Kernel Sizes of CMST Blocks.} The CMST blocks employ varying kernel sizes along the temporal axis, while keeping the spatial kernel size fixed at 3.}\label{tab:kernel_size}
    \centering
    \begin{tabular}{{@{}c|c|c@{}}} 
        \toprule
        \textbf{CMST Level} & \textbf{Temporal} & \textbf{Spatio} \\
        \midrule
        1                   & 1, 25             & 3               \\
        2                   & 1, 13, 21         & 3               \\
        3                   & 1, 7, 9           & 3               \\
        4                   & 1, 3, 5           & 3               \\
        \bottomrule
    \end{tabular}
\end{table}

\begin{figure}[!t]
    \centering
    \includegraphics[width=1.0\textwidth]{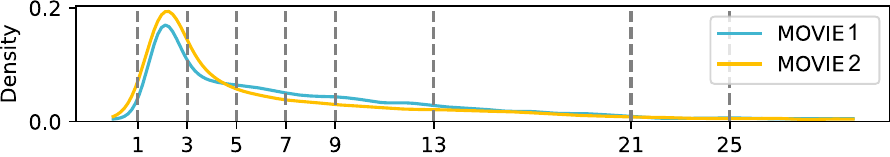}
    \caption{
        \textbf{RGC durations.} The grey line represents the high-frequency durations of RGCs, where we manually identified several peaks. For entirely including a complete duration, we choose a truncation position slightly lagging behind the peak.
    }\label{fig:RGC_durations}
\end{figure}

These kernel sizes are chosen by considering the duration of the neural responses. We calculated the density function of entire neural responses' duration and identified several high-frequency durations, which we use as feature-sensitive spans, used for the kernel sizes of CMST blocks. Here, durations refer to the durations of firing rates ranging from 0 to maximum and then back to 0. We estimated their distribution using kernel density estimation. We manually identified several peaks, as shown in \cref{fig:RGC_durations}, where RGC durations are different in two different video stimuli.

Finally, the latent variables are flattened and passed through a linear layer to obtain the prediction results for RGCs. In this study, we set the number of RGCs to 90, which is predicted simultaneously as a population coding scheme.

\subsection{Loss Function}

Initially, we employed Root Mean Square Error (RMSE) as the loss function, at \cref{eq:rmse}. However, experimental results indicated that using RMSE as the target did not adequately consider temporal dynamics, resulting in lower generalization capabilities of the model.

\begin{equation}
    \mathcal{L_{\text{RMSE}}} = \sqrt{\frac{1}{n}\sum_{i=1}^{n}(y_i - \hat{y}_i)^2} 
    \label{eq:rmse}
\end{equation}

Therefore, we introduced a Soft Dynamic Time Warping (SoftDTW) loss function~\cite{cuturi_soft-dtw_2018}. The SoftDTW, unlike Euclidean losses such as RMSE, considers potential time shifts or variations of length of durations. It is notable that using SoftDTW for predicting longer time windows may lead to distortion and difficulty in representing local abrupt changes. Thus, we group the time sequences into shorter subsequences in the rolling window way and then compute as \cref{eq:softdtw}:

\begin{equation}
    \mathcal{L_{\text{SoftDTW}}}^n = \frac{1}{L-n}\sum_{i=1}^{L-n}\text{SoftDTW}(y_i, \hat{y}_i), i \in \{1,2, \dots, L-n\} 
    \label{eq:softdtw}
\end{equation}

Here, $n$ demonstrates the length of the subsequence, which is set to 6 or 12, and $L$ is the length of the time sequence. Meanwhile, to reduce the model's predictions of meaningless negative responses, we add a negative ReLU function as a penalty term to the loss function, as shown in \cref{eq:relu}.

\begin{equation}
    \mathcal{L_{\text{-ReLU}}} = \frac{1}{n}\sum_{i=1}^{n}\max(0, -\hat{y}_i) 
    \label{eq:relu}
\end{equation}

Then, the entire loss function is represented as shown in \cref{eq:vist_loss}, named Vi-ST loss in our model.

\begin{equation}
    \mathcal{L_{\text{Vi-ST}}} =
    \alpha\mathcal{L_{\text{RMSE}}} +
    \beta\mathcal{L_{\text{-ReLU}}} +
    \gamma\mathcal{L_{\text{SoftDTW}}}^{6} +
    \gamma\mathcal{L_{\text{SoftDTW}}}^{12}
    \label{eq:vist_loss}
\end{equation}

Here, $\alpha$, $\beta$, and $\gamma$ are hyperparameters, and we set them at 0.1, 0.5, and 5$\times$10$^{-6}$, respectively.

\section{Experiments}
\subsection{Dataset}
\subsubsection{RGCs Responses}

This work focuses on studying the retina encoding mechanism by recording RGC spikes in response to visual stimuli. The dataset used was organized by \cite{onken_using_2016}, and has been used in various studies on retinal coding\cite{zheng_unraveling_2021, wangSpikeSEEEnergyefficientDynamic2023, yan_revealing_2022, NEURIPS2023_0bcf9cf6, liFusionANNsDecoder2022, zhangReconstructionNaturalVisual2020}. A total of 90 well-responding RGCs were used for training and testing. In summary, the experimental setup involves isolating the retina from salamanders, inverting the ganglion cell layer downward onto a multielectrode array, stimulating the photoreceptor layer of the retina using an OLED display, and recording the responses of RGCs. For each cell, receptive fields were determined by computing the spike-triggered average using stimulus images of spatiotemporal white noise \cite{chichilniskySimpleWhiteNoise2001}. Singular value decomposition was used to separate the spike-triggered average into a spatial and temporal component \cite{gauthierReceptiveFieldsPrimate2009}. Finally, a two-dimensional Gaussian function was fitted to the spatial receptive field component to determine the center, size, and shape of the receptive field. Receptive fields were normalized to unit Euclidean norm. The details of the experimental procedures and data analysis are standard operations for retinal coding studies \cite{onken_using_2016, liuInferenceNeuronalFunctional2017, liuSimpleModelEncoding2022}.

To align frame rates, neuronal spikes were grouped into spike counts at intervals of 33 ms, resulting in a dataset with 30 Hz response for each trial, in line with the sampling rate of 30 Hz used by video stimuli. In other words, spike counts within each 33 ms interval are recorded as the response data for that time period. To reduce noise during the recording process, several trials were conducted for the same stimulus, and response data was averaged over the trials to obtain the neuronal firing rate for each cell and each video.

\subsubsection{Natural Scene Stimuli}

The natural scene stimuli used in this study consist of two video clips with rich natural scenes as shown in \cite{onken_using_2016}, named \textit{Mov1} and \textit{Mov2}.~\textit{Mov1} features minimal changes between video frames and mainly captures salamanders swimming in water, while \textit{Mov2} exhibits more significant changes between video frames and mainly documents tigers and deers in outdoor activities. Both video clips have a resolution of 360$\times$360 pixels with a frame rate of 30 fps, with \textit{Mov1} being 60 seconds long (1800 frames) and \textit{Mov2} being 53.3 seconds long (1600 frames). The original video clips were in RGB color with three channels. To reduce the impact of color information on visual stimuli, video clips were converted to grayscale image stacks with brightness values ranging from 0 to 255. Details of video information can be found in \cite{onken_using_2016}. These two videos show different levels of scene complexity in both spatial and temporal domains, which presents a challenge for the building and testing of encoding models \cite{zheng_unraveling_2021}.

\subsection{Metrics}

In the task of neural encoding, the Pearson correlation coefficient (CC) is commonly used to assess the reliability of predicting neural signals~\cite{zheng_unraveling_2021,Wang2023.03.21.533548,turishcheva2023dynamic}, and it is not specifically stated that the mean value of CC is used as the metric. While CC considers the macro trends of the entire sequence, it lacks the detailed consideration of temporal information or dynamics over time. In line with the departure point of the Vi-ST loss, we attempt to introduce the factor of spike duration to observe the neuronal encoding performance of the model.

This metric initially selects the lengths of corresponding subsequences in the response sequence, representing the duration of a complete neural response (from non-spike to non-spike). In addition, the lengths are used to estimate a continuous density function using Gaussian kernel density estimation. Sampling is then performed in the linear space formed by the upper and lower bounds of the set of lengths of target and predicted subsequences. Finally, the Kullback-Leibler (KL) divergence from the predicted distribution to the target distribution is computed, clipped between 0 and 1000. We refer to this metric as SD-KL. The detailed algorithm process is provided in the Supplementary Materials.

\subsection{Implementation Details}

\subsubsection{Pre-trained ViT}

Oquab \etal\cite{oquab_dinov2_2023} contributed to the open-source DINOv2 model. Additionally, Darcet \etal\cite{darcet_vision_2023} proposed Registers Token, enhancing the robustness of DINOv2. Considering computational costs, we chose the \textit{ViT-L/14} with registers, obtained through \textit{ViT-g/14} distillation as the ViT Prior. After feeding images into DINOv2, the preprocessing produces feature maps of size 16$\times$16, a process referred to as \textit{patchfy}. Each patch corresponds to a 14$\times$14 region after scaling down from the original image~\cite{oquab_dinov2_2023}.

\subsubsection{RGC Receptive Field}

The RF, with an initial size of 90$\times$90, is downsampled to 16$\times$16 to match the feature map size of DINOv2. Initially, we directly added the RF on feature latent variables from the previous module. However, experimental results show that the 3D AdaLN Zero module offers better robustness.

\subsubsection{Training}

To maximize the generalization ability of the model, we trained and tested the model using data from different videos, such as training on Mov1 and validating on Mov2, marked as \textit{Mov1$\rightarrow$Mov2}, and vice versa. Due to computational constraints, we apply the rolling window to crop 128-frame clips as training samples. Each epoch involved arbitrarily selecting 768 samples for the training set with a batch size of 16. The validation set uses the entire video as a sample.

For optimization, we choose AdamW~\cite{loshchilov_decoupled_2019} with a weight decay of 0.1. We trained for 30 epochs, applying linear learning rate warm-up to 8$\times$10$^{-4}$ for the first 5 epochs, then gradually decaying to 1$\times$10$^{-4}$ using a half-cycle cosine schedule. At the last 10 epochs, Stochastic Weight Averaging~\cite{izmailov_averaging_2019} was applied to mitigate model overfitting.

\subsection{Results}
\begin{table}[!tbh]
    \caption{
        \textbf{Comparison with Previous Works.} Comparison of the mean CC of Vi-ST with other models on the same dataset.~\textit{Mov1$\rightarrow$Mov1} and \textit{Mov2$\rightarrow$Mov2} represent training and testing on the same video, while \textit{Mov1$\rightarrow$Mov2} and \textit{Mov2$\rightarrow$Mov1} represent training and testing on different videos. Notably, CRNN selects 80 RGCs, while we use 90 RGCs, and we introduced regularization methods for avoiding over-fitting. These may introduce some errors. Using RGC data to predict across-movie data is quite different from within-movie tasks, and the within-movie metrics are only for reference. Our focus in this study is to obtain better generalization ability across different videos with the same model.}\label{tab:comparison_previous_works}
    \centering
    \begin{tabular}{{@{}c|c|c@{}}} 
        \toprule
        \textbf{Model}                   & \textbf{\textit{Mov1$\rightarrow$Mov1}} & \textbf{\textit{Mov2$\rightarrow$Mov2}} \\
        \midrule
        CRNN\cite{zheng_unraveling_2021} & \textbf{0.857}                          & \textbf{0.718}                          \\
        I3D+MSTCN                        & 0.846                                   & 0.668                                   \\
        DINOv2+MSTCN                     & 0.849                                   & 0.672                                   \\
        Vi-ST                            & 0.789                                   & 0.570                                   \\
        \bottomrule
        \textbf{Model}                   & \textbf{\textit{Mov1$\rightarrow$Mov2}} & \textbf{\textit{Mov2$\rightarrow$Mov1}} \\
        \midrule
        I3D+MSTCN                        & 0.108                                   & 0.074                                   \\
        DINOv2+MSTCN                     & 0.101                                   & 0.100                                   \\
        Vi-ST                            & \textbf{0.334}                          & \textbf{0.281}                          \\
        \bottomrule
    \end{tabular}
\end{table}

\begin{figure}[!th]
    \centering
    \includegraphics[width=1.0\textwidth]{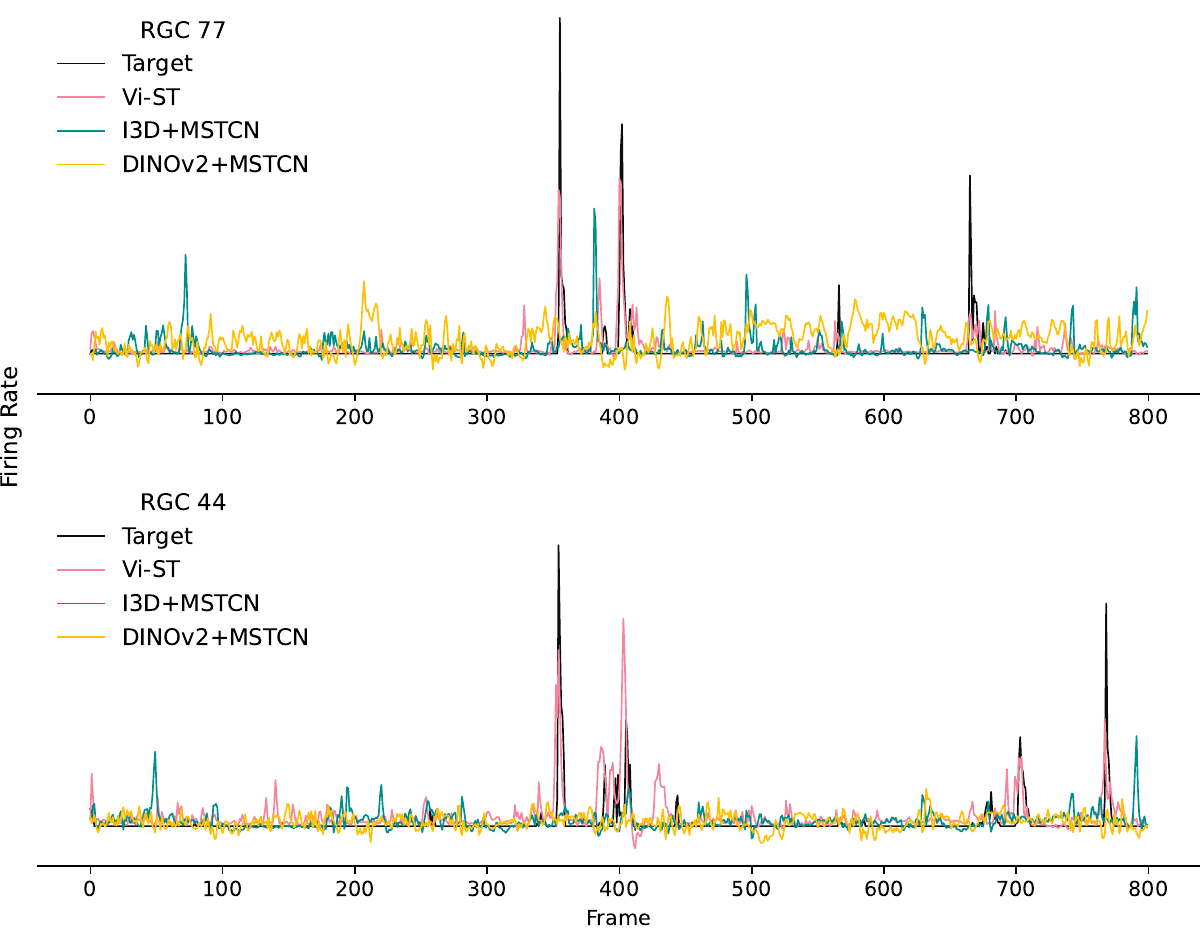}
    \caption{
        \textbf{RGC Prediction Results.}
        We visualized two example cells showing representative prediction results. By comparing the response
        results of different models, we can intuitively understand the relationship between the metric and the actual response. More examples can be found in the Supplementary Materials. }\label{fig:real_predictions}
\end{figure}

\subsubsection{Comparison with Previous Works}

Previous works mainly focus on training and testing the data from the same video~\cite{zheng_unraveling_2021, wangSpikeSEEEnergyefficientDynamic2023, NEURIPS2023_0bcf9cf6, liFusionANNsDecoder2022}. Typical CC metrics on these two videos are shown in \cref{tab:comparison_previous_works}. However, our focus places a significant emphasis on spatiotemporal modeling tailored for neural encoding tasks across different videos. To ensure a fair comparison of model performance, we constructed two control models. The first employs an I3D feature extractor as a backbone and MSTCN as the downstream model, denoted as \textit{I3D+MSTCN}; The second model utilizes DINOv2 as the prior and MSTCN as the downstream model, labeled as \textit{DINOv2+MSTCN}.
Performance in training and testing using the same videos shows relatively good CCs; however, a significant improvement in transfer prediction between videos can only be observed with Vi-ST, and other models lack the generalization ability as shown in \cref{tab:comparison_previous_works}. Looking into the details of single neuron performance, we found that Vi-ST demonstrates a better ability to predict spare neuronal responses, as shown in \cref{fig:real_predictions}.

\begin{figure}[tbh]
    \centering
    \begin{subfigure}{0.48\linewidth}
        \includegraphics[width=1.0\linewidth]{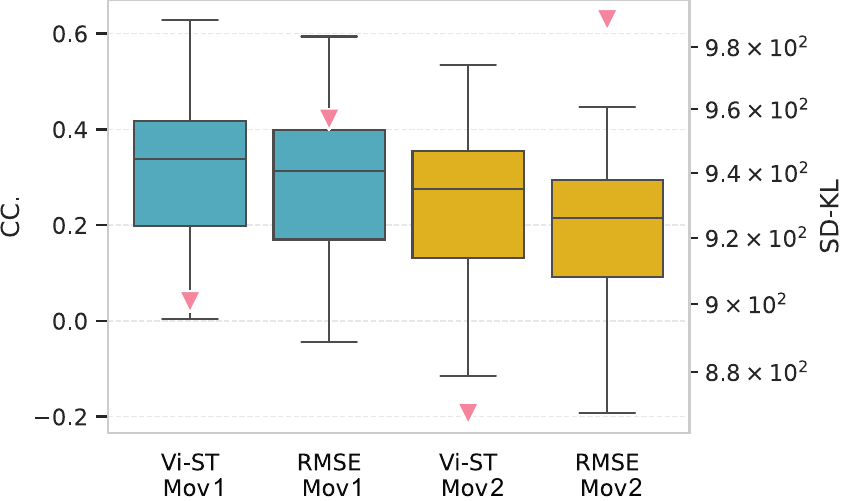}\caption{}\label{fig:loss_comparison}
    \end{subfigure}
    \hfill
    \begin{subfigure}{0.48\linewidth}
        \includegraphics[width=1.0\linewidth]{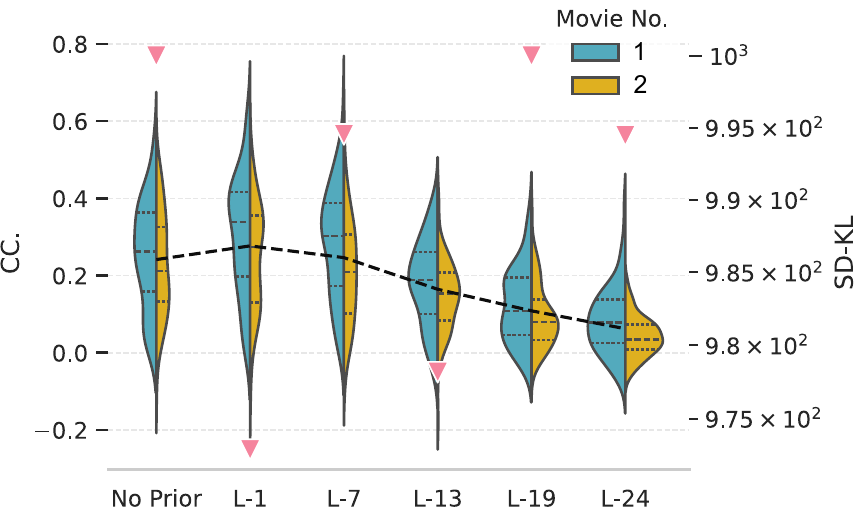}\caption{}\label{fig:dinov2_prior}
    \end{subfigure}
    \caption{
        \textbf{Loss Function Comparison and DINOv2 Prior Comparison} with CC (left y-axis) and SD-KL (right y-axis, representing by red triangles) (a) The comparison of Vi-ST loss and RMSE loss, where \textit{Mov1} represents training on \textit{Mov1} and testing on \textit{Mov2}, and vice versa.\ (b) The comparison of different DINOv2 priors, where the layer number of DINOv2 are represented as $L-n$, \eg~$L-1$ represents the first layer.}\label{fig:loss_dinov2}
\end{figure}

\subsubsection{Comparison of Loss Functions}
We compared the model performance with RMSE loss and Vi-ST loss. In \cref{fig:loss_comparison}, we observe that the Vi-ST loss outperforms RMSE in all experiments. This indicates that considering temporal variations in the loss function is crucial for neural encoding tasks.

\begin{figure}[!th]
    \centering
    \begin{subfigure}{0.48\linewidth}
        \includegraphics[width=1.0\linewidth]{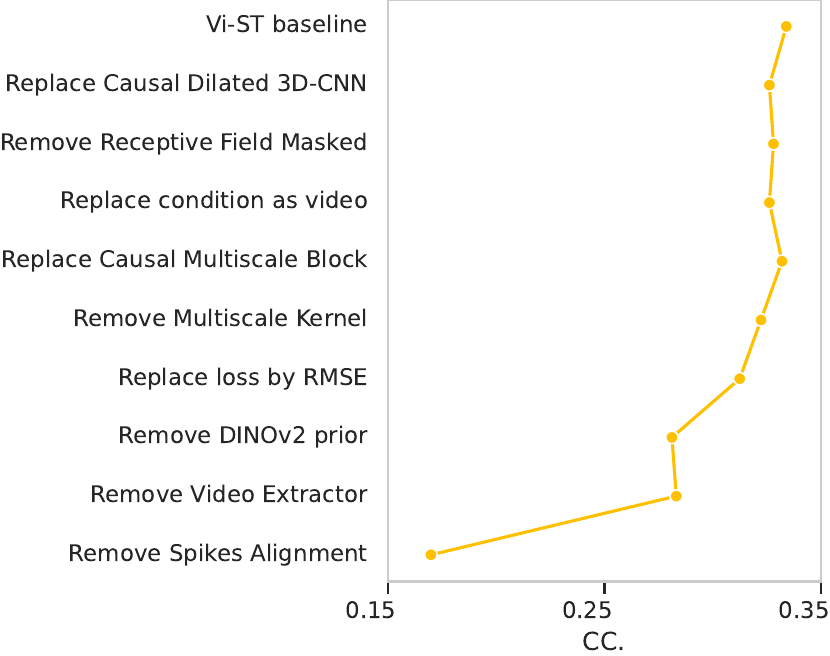}\caption{}\label{fig:ablation}
    \end{subfigure}
    \hfill
    \begin{subfigure}{0.48\linewidth}
        \includegraphics[width=1.0\linewidth]{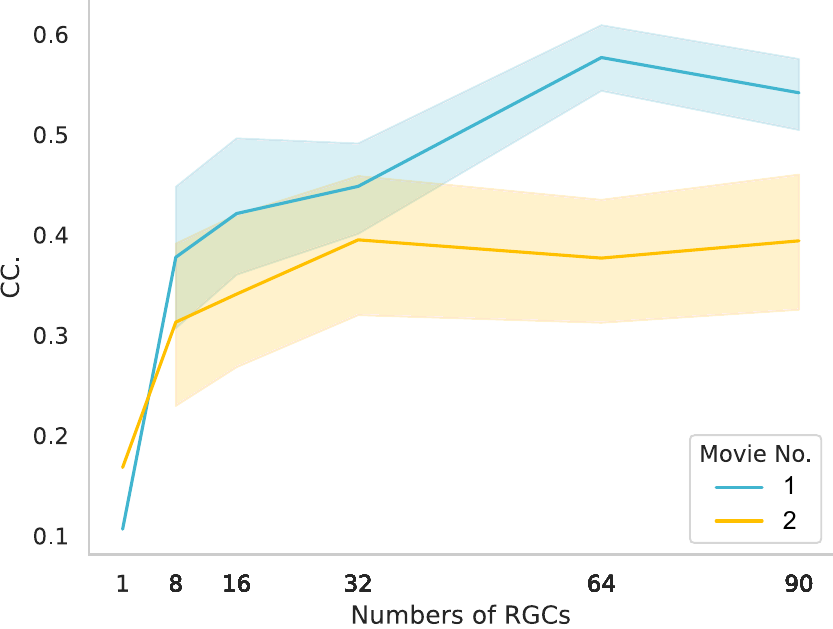}\caption{}\label{fig:complementary_coding}
    \end{subfigure}
    \caption{
        \textbf{Ablation Study and Complementary Coding.}
        (a) The ablation study of Vi-ST model by removing or replacing different modules.
        (b) Comparison of the benefits of complementary coding as the number of encoded neurons increases.
    }\label{fig:ablation_complementary_coding}
\end{figure}

\subsubsection{Comparison of ViT Prior}
Amir \etal\cite{amir_effectiveness_2023} points out that the hierarchical structure of ViT provides different representations, where the earlier layers contain lower semantic information but better present positional information. Therefore, we use inputs from different layers of DINOv2 as priors, specifically, the 1st, 7th, 13th, 19th, and the last layer, along with using only the patchfy representations as a prior, to test the performance of Vi-ST.~In \cref{fig:dinov2_prior}, we observe that the representations from the earlier layers of ViT are beneficial for neural encoding results, aligning with the biological characteristics of the retina.

\subsubsection{Ablation study}

To investigate the contribution of each model module to prediction performance, we progressively removed or replaced different modules and observed changes in CC.~We found that the Spikes Alignment module played a more significant role compared to the Video Extractor, and the Vi-ST loss had a non-negligible impact on achieving more generalized neural encoding as well. Additionally, in conjunction with the results in \cref{fig:ablation}, we observed that removing the ViT prior had a significant impact on the model, and excessively deep ViT features did not bring about more gains. Therefore, selecting an appropriate spatial feature prior for RGC visual encoding tasks requires careful consideration.

\subsubsection{Comparison of benefits of complementary coding}
The prediction results for neural encoding tasks often do not involve a fixed number of neurons\cite{consortium_functional_2021,wang_towards_2023}. Therefore, we ask: \textit{Is it optimal to construct an end-to-end model capable of simultaneously predicting all neural responses?}

Ding \etal\cite{ding_information_2023} explored the benefits of complementary encoding for neural encoding by constructing a high-dimensional representation manifold of responses of the RGC population. This suggests that collaboration between neurons can compensate for the problem of insufficient expression when the firing rate saturates. We conducted experiments to validate this conclusion. We sorted the results predicted by Vi-ST and selected the top eight RGCs based on their CC.~We then limited the number of RGCs predicted by Vi-ST to 90, 64, 32, 16, 8, and 1, respectively. In \cref{fig:complementary_coding}, we observe that the CC stabilizes as the number of encoded neurons increases. This implies that, similar to biological RGC responses, a redundant encoding space allows Vi-ST to better match real response outcomes.

\section{Conclusion}
In this work, we proposed the Vi-ST model to align dynamic visual stimuli of the natural scene with RGC responses. The experiments demonstrate that the Vi-ST model can learn certain encoding mechanisms akin to biological visual neural coding. Through ablation experiments, we identified that the primary capabilities of the Vi-ST model stem from the Spike Alignment module and the inclusion of the ViT prior. By contrasting the benefits of complementary coding, we believe that the model has potential applications in a wider range of neural encoding tasks. Additionally, we are exploring new metrics to examine the model performance in representing target temporal dynamics.

This study presents certain limitations that need further investigation. Firstly, the SD-KL metric exhibits sensitivity to noise and demonstrates limited expressive power. Additionally, the robustness of Vi-ST needs to be confirmed through the acquisition of more extensive datasets, including a greater number of recorded cells and a variety of video types. Future research will aim to refine and overcome these challenges by leveraging recent advancements in experimental techniques. Furthermore, we plan to extend our approach to the study of visual coding using neurons from other parts of visual systems, such as the visual cortex.


\section*{Acknowledgements}
We thank Shanshan Jia, Zhile Yang, Zerui Yang and Jing Peng for the highly valuable discussions.

%
%
\bibliographystyle{splncs04}
\bibliography{main}
\end{document}